\documentclass[10pt,twocolumn,letterpaper]{article}

\usepackage[pagenumbers]{iccv} 

\usepackage{newtxtext}
\definecolor{iccvblue}{rgb}{0.21,0.49,0.74}
\usepackage[pagebackref,breaklinks,colorlinks,allcolors=iccvblue]{hyperref}

\def\paperID{6096} 
\def\confName{ICCV}
\def\confYear{2025}

\title{TemCoCo: Temporally Consistent Multi-modal Video Fusion with Visual-Semantic Collaboration}

\author{
    Meiqi Gong \quad
    Hao Zhang\footnotemark[1] \quad
    Xunpeng Yi \quad
    Linfeng Tang \quad
    Jiayi Ma\footnotemark[1] \footnotetext[1]{\thanks{Corresponding author.}}\\
    Electronic Information School, Wuhan University, Wuhan 430072, China \quad \\
    {\tt\small  \{meiqigong, yixunpeng\}@whu.edu.cn, \{zhpersonalbox, linfeng0419, jyma2010\}@gmail.com}
}

\begin{document}
\maketitle
\begin{abstract}
    Existing multi-modal fusion methods typically apply static frame-based image fusion techniques directly to video fusion tasks, neglecting inherent temporal dependencies and leading to inconsistent results across frames. To address this limitation, we propose the first video fusion framework that explicitly incorporates temporal modeling with visual-semantic collaboration to simultaneously ensure visual fidelity, semantic accuracy, and temporal consistency. First, we introduce a visual-semantic interaction module consisting of a semantic branch and a visual branch, with Dinov2 and VGG19 employed for targeted distillation, allowing simultaneous enhancement of both the visual and semantic representations. Second, we pioneer integrate the video degradation enhancement task into the video fusion pipeline by constructing a temporal cooperative module, which leverages temporal dependencies to facilitate weak information recovery. Third, to ensure temporal consistency, we embed a temporal-enhanced mechanism into the network and devise a temporal loss to guide the optimization process. Finally, we introduce two innovative evaluation metrics tailored for video fusion, aimed at assessing the temporal consistency of the generated fused videos. Extensive experimental results on public video datasets demonstrate the superiority of our method. Our code is released at \url{https://github.com/Meiqi-Gong/TemCoCo}.
\end{abstract}
\section{Introduction}
\label{sec:intro}

\begin{figure}[t]
  \centering
   \includegraphics[width=1\linewidth]{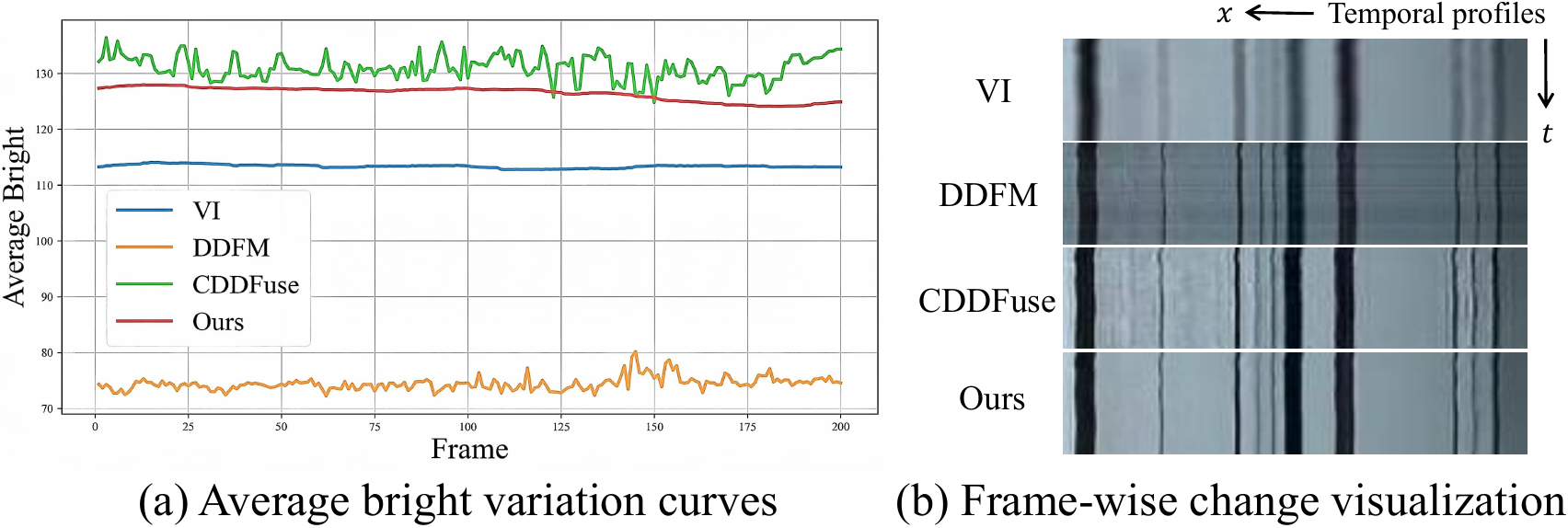}
   \caption{Comparison of temporal consistency between two frame-by-frame fusion methods (DDFM and CDDFuse) and our TemCoCo for video fusion. (a): The average brightness variation curves for continuous 200 frames. (b): Visual effects of the same row in continuous 50 frames.}
   \label{fig:intro_example}
\end{figure}

The limitations of imaging principles inherently constrain the ability of a single modality to provide a comprehensive representation of a scene~\cite{gong2024cross,cao2023multi,li2024dstcfuse}. For instance, whether infrared or visible, each modality has its unique strengths and weaknesses~\cite{ma2019infrared, zhang2021image}. By leveraging the complementary characteristics of these modalities, multi-modal image fusion enables a more thorough and accurate depiction of the scene, addressing the inherent limitations of individual modalities~\cite{xu2022cufd, fu2023lraf}. Owing to these characteristics, fusion images are more beneficial for practical downstream tasks such as autonomous driving~\cite{cui2021deep,liu2023multi}, security surveillance~\cite{liu2024asifusion,liu2024coconet}, and others~\cite{luo2022thermal,qi2023generative,liu2025infrared}.

Although significant progress has been made in static frame-based fusion, real-world dynamic environments demand continuous video sequences as the perceptual foundation~\cite{chan2021basicvsr, yang2024motion}. In such scenarios, preserving spatial fidelity while ensuring temporal consistency becomes essential~\cite{voeikov2020ttnet,wang2022adafocusv3}. Compared to static frame-based methods, video-based approaches integrate spatiotemporal features, making them more suitable for handling dynamic scene challenges like object motion and environmental changes~\cite{liu2022video, xiong2010robust}, and thus more applicable to real-world dynamic scenes.

Current research remains predominantly focused on static frame-based image fusion approaches. The rapid evolution of deep learning has revolutionized the paradigm of multi-modal image fusion, marking a significant shift from traditional methods to sophisticated deep learning-based methods~\cite{zhang2023visible,zhang2021deep,xu2023dm}. For instance, CrossFuse~\cite{li2024crossfuse} focuses on mining complementary information between multi-modality image frames by introducing a cross-attention mechanism that captures spatially complementary cues within each frame. SwinFusion~\cite{ma2022swinfusion} introduces a transformer-based framework to enhance fusion by modeling long-range spatial dependencies and capturing both intra-modal and cross-modal interactions within single-frame inputs. Besides, GAN-based and diffusion-based approaches model fusion from distribution perspectives. GANMcC~\cite{ma2020ganmcc} introduces a multi-classification constrained GAN framework, which achieves deep infrared-visible fusion by modeling spatial feature distributions of single-frame inputs. Diff-IF~\cite{yi2024diff} establishes a diffusion paradigm that performs denoising iterations to enable feature interaction within the single-frame spatial domain, progressively coupling two modalities to discover an optimized fusion distribution. Recently, there has been growing awareness of the significance of continuous video fusion~\cite{xie2024rcvs,tang2024infrared}. Although several methods have made preliminary attempts to explore video fusion, they fail to modeling temporal dependencies. For example, DFTP applys varying fusion rules across frames to account for temporal fluctuations, rather than leveraging the temporal structure inherent in video data~\cite{guo2024dftp}. Consequently, current research on video fusion remains in its infancy, lacking frameworks capable of jointly capturing spatial and temporal dynamics.


Overall, existing methods face challenges in achieving high fusion quality and temporal consistency. These approaches fail to guide and couple visual information with semantic content in a robust manner, thereby compromising semantic accuracy and visual fidelity, which limits subsequent applications, whether human judgment or machine inference~\cite{tang2022image}. Furthermore, current methods typically treat video fusion as a direct extension of frame-by-frame image fusion~\cite{xie2024rcvs}, neglecting the temporal dependencies between frames, thus exhibiting temporal inconsistency and visible artifacts~\cite{chan2021basicvsr}. To substantiate these claims, Figure~\ref{fig:intro_example} presents two state-of-the-art static frame-based fusion methods CDDFuse~\cite{zhao2023cddfuse} and DDFM~\cite{zhao2023ddfm}, using both quantitative and qualitative evaluations. As shown in Figure~\ref{fig:intro_example}(a), CDDFuse and DDFM exhibit significant fluctuations in the average brightness variation. In addition, the frame-wise change visualizations in Figure~\ref{fig:intro_example}(b) further reveal the temporal instability of these methods.


To address the aforementioned challenges, we propose \textbf{TemCoCo}, the first temporally-consistent multi-modal video fusion framework that explicitly incorporates temporal modeling and visual-semantic collaboration. TemCoCo is designed to simultaneously capture temporal dynamics and enriched visual-semantic features from degraded source videos, significantly enhancing visual fidelity, semantic accuracy, and temporal consistency. Firstly, we design a novel visual-semantic interaction module comprising a visual branch and a semantic branch. By leveraging the powerful prior knowledge of foundation models (Dinov2~\cite{oquab2023dinov2} and VGG19), it enhances representational capacity of each branch and ensures meaningful guidance, effectively balancing semantic richness and visual fidelity. To the best of our knowledge, this is the first work to explicitly integrate semantic guidance into video fusion. Secondly, we pioneer the unification of video enhancement and video fusion within a single framework. A temporal cooperative module leverages inter-frame dependencies to recover weak information, while a data augmentation strategy simulates common degradations (\emph{e.g.}, noise and blur) to enhance robustness. This represents the first attempt to jointly address video degradation and video fusion. Thirdly, to enhance temporal consistency, we introduce a temporal-enhanced mechanism coupled with a dedicated temporal consistency loss. The mechanism, comprising 3D convolutions and temporal attention, explicitly models both local and global inter-frame dependencies to capture dynamic scene variations. Meanwhile, the temporal loss enforces temporal alignment by matching frame-wise variations between the fused and source videos. This synergistic combination offers a novel solution for achieving high temporal consistency in results. As illustrated in Figure~\ref{fig:intro_example}, our proposed TemCoCo yields smoother brightness transitions and higher temporal consistency results than CDDFuse and DDFM, reflecting the effectiveness of our temporal design. Finally, we propose two novel metrics, feaCD and flowD, specifically designed to evaluate temporal coherence in fused videos. feaCD assesses feature-level consistency by measuring the direction of high-level representations across frames, while flowD quantifies image-level smoothness by analyzing optical flow variations.

In summary, our contributions are as follows:
\begin{itemize}
  \item We propose the first multi-modal video fusion framework that explicitly incorporates both temporal modeling and visual-semantic collaboration, enhancing visual fidelity, semantic accuracy, and temporal consistency.
	\item We pioneer the unification of video enhancement with video fusion through temporal cooperative modules and targeted guidance, enabling the network to learn robust feature representations in a unified framework.
	\item We introduce a temporal-enhanced mechanism and a temporal loss, which jointly capture local and global inter-frame dependencies, ensuring smooth and coherent transitions in the fused video.
	\item We propose two novel temporal metrics to evaluate the temporal consistency between generated and original videos at the feature and image levels, respectively.
\end{itemize}

\begin{figure*}[t]
  \centering
   \includegraphics[width=0.98\linewidth]{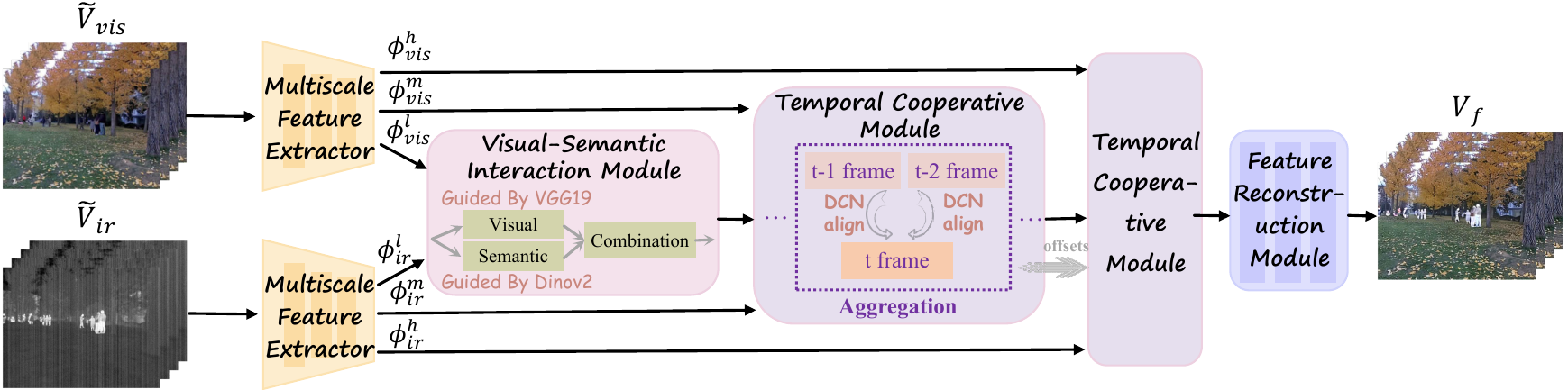}
   \vspace{-0.1in}
   \caption{The overall framework of the proposed TemCoCo.}
   \label{fig:overview}
\end{figure*} 

\section{Method}

In this section, we provide a detailed description of our proposed method, TemCoCo, including its overall framework, design details and loss functions.

\subsection{Overview}

\begin{figure*}[t]
    \centering
        \includegraphics[width=1\linewidth]{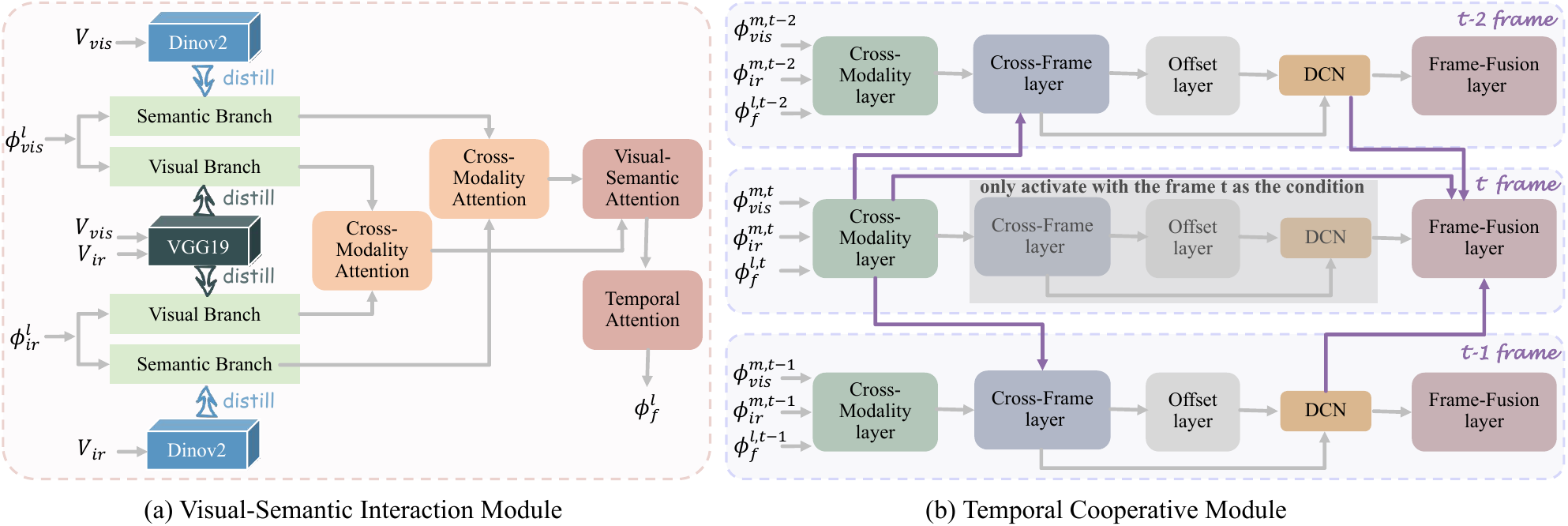}
        \vspace{-0.2in}
        \caption{Architectures of (a) Visual-Semantic Interaction Module and (b) Temporal Cooperative Module.}
        \label{fig:details}
\end{figure*}

Given a degraded visible video $\tilde{V}_{vis}\in\mathbb{R}^{T\times H \times W \times 3}$ and its corresponding degraded infrared video $\tilde{V}_{ir}\in\mathbb{R}^{T\times H \times W \times 1}$, where $T,H$ and $W$ represent the temporal length and spatial dimensions respectively, our objective is to generate a fused video $V_f\in\mathbb{R}^{T\times H \times W \times 3}$ that maintains high fidelity in overall representation, demonstrates superior semantic accuracy, and preserves temporal consistency across frames. We present the overall framework in Figure~\ref{fig:overview}. To achieve spatio-temporal interaction during the fusion process, we activate the temporal-enhanced mechanism to replace the conventional 2D convolutions throughout the network with 3D convolutions. Initially, both $\tilde{V}_{vis}$ and $\tilde{V}_{ir}$ are processed through a Multi-Scale Feature Extractor to obtain hierarchical representations at three distinct scales:
\begin{align}
    \phi^h_{vis},\phi^m_{vis},\phi^l_{vis}&=E_{vis}(\tilde{V}_{vis}),\\
    \phi^h_{ir},\phi^m_{ir},\phi^l_{ir}&=E_{ir}(\tilde{V}_{ir}),
\end{align}
where $E^{vis}$ and $E^{ir}$ denote the feature extraction networks for visible and infrared modalities, respectively, and superscripts $h$, $m$, and $l$ representing high-, medium-, and low-scale features. The low-scale features $\phi^l_{vis}, \phi^l_{ir}$ are fed into the Visual-Semantic Interaction Module (VSIM) to achieve the separation and aggregation of semantic and visual features. Through targeted guidance from the foundation model, this process enhances both visual quality and semantic accuracy. Subsequently, the medium- and high-scale features are processed by two progressive Temporal Cooperative Modules (TCM), which leverage inter-frame correlations to enhance feature representation. Finally, the enhanced features are reconstructed into the video space through the feature reconstruction module:
\begin{equation}
 V_f=RM(\phi_f^h),
\end{equation}
where $RM$ denotes the feature reconstruction network and $\phi_f^h$ represents the fused feature. In the following, we will provide detailed descriptions of the VSIM and TCM.

\subsection{Visual-Semantic Interaction Module (VSIM)}
VSIM is dedicated to extracting enhanced visual-semantic features, with its overall process illustrated in Figure~\ref{fig:details}(a). First, the module processes low-scale features through dual parallel branches for specialized visual and semantic feature extraction through:
\begin{align}
    \phi^s_{vis},\phi^s_{ir}&=SeB(\phi^l_{vis},\phi^l_{ir}),\\
    \phi^v_{vis},\phi^v_{ir}&=ViB(\phi^l_{vis},\phi^l_{ir}),
\end{align}
where $SeB$ and $ViB$ represent the semantic branch network and the visual branch network, respectively. To achieve domain-specific feature enhancement, we strategically integrate VGG19 for visual feature refinement and Dinov2 for semantic understanding, leveraging their powerful priors through knowledge distillation. This integration facilitates the extraction of robust features while preserving domain-specific characteristics. Notably, the input to the foundation models is the enhanced original video, enabling them to assist in recovering features from weak information. Subsequently, the outputs from both modalities through the semantic branch undergo Cross-Modality Attention (CM-Attn) based on SwinTransformer~\cite{liu2021swin} to enhance inter-modal semantic information interaction:
\begin{equation}
    \begin{aligned}
 \text{CM-Attn}(\phi^x_{vis},\phi^x_{ir}) =
 \left\{ \text{softmax}\Big(\frac{Q_{vis}K_{ir}}{\sqrt{d_k}}\Big)V_{ir}, \right. \\
 \left. \text{softmax}\Big(\frac{Q_{ir}K_{vis}}{\sqrt{d_k}}\Big)V_{vis} \right\}, x\in\{v,s\},
    \end{aligned}
\end{equation}
where $Q\in\mathbb{R}^{\frac{HW}{16} \times C}$, $K\in\mathbb{R}^{\frac{HW}{16} \times C}$ and $V\in\mathbb{R}^{\frac{HW}{16} \times C}$ represent the query, key, and value matrices, respectively. The query $Q$ is alternately provided by the visible and infrared modalities, while the key $K$ and value $V$ are supplied by the other modality. The temporal dimension $T$ has been merged into the batch dimension. After that, the cross-modality enhanced features are processed through Semantic-Visual Attention (SV-Attn) to facilitate semantic and visual bidirectional feature interaction. Similar to CM-Attn, the visual and semantic features alternately serve as the query ($Q$) and key-value pairs ($KV$). The SV-Attn can be formulated as:
\begin{equation}
    \begin{aligned}
 \text{SV-Attn}(\!\phi^v\!,\!\phi^s\!)\! =\!
 \left\{\! \text{softmax}\!\Big(\!\frac{Q\!_{v}\!K\!_{s}}{\sqrt{d_k}}\!\Big)\!V\!_{s}, \text{softmax}\!\Big(\!\frac{Q\!_{s}\!K\!_{v}}{\sqrt{d_k}}\!\Big)\!V\!_{v} \!\right\}\!.
    \end{aligned}
\end{equation}
Finally, temporal attention (T-Attn) is applied by flattening features along temporal and channel dimensions to obtain $Q, K, V\in\mathbb{R}^{T \times C}$:
\begin{equation}
 \text{T-Attn}(\phi) = \text{softmax}\Big(\frac{Q_{t}K_{t}}{\sqrt{d_k}}\Big)V_{t}.
\end{equation}

\subsection{Temporal Cooperative Module (TCM)}
TCM facilitates weak information recovery by leveraging inter-frame correlations, with its detailed architecture presented in Figure~\ref{fig:details}(b). The diagram specifically demonstrates the processing pipeline for the t-th frame $\phi_f^{l,t}$, while other frames follow an analogous procedure. Notably, the first two frames of the video sequence undergo self-referential enhancement without inter-frame assistance. TCM comprises five fundamental components: cross-modality layer, inter-frame layer, offset layer, DCN~\cite{zhu2019deformable}, and frame fusion layer. The processing pipeline begins with the cross-modality layer receiving medium-scale features $\phi_{vis}^m$ and $\phi_{ir}^m$ from the encoder, along with temporally enhanced features $\phi_{f}^l$ from VSIM as:
\begin{equation}
    \phi_{f}=CML(\phi_{vis}^m,\phi_{ir}^m,up(\phi_{f}^l)),
\end{equation}
where $CML$ denotes  the cross-modality layer network and $up$ is the upsampling operation. For the $t$-th frame $\phi_f^{t}$, we select its two preceding frames $\phi_f^{t-1}$ and $\phi_f^{t-2}$ as conditional references to assist in weak information recovery. The inter-frame layer and offset layer learns the spatial offsets $o_1$ and $o_2$ from $\phi_f^{t-1}$ and $\phi_f^{t-2}$ to $\phi_f^{t}$, enabling frame alignment through DCN, formulated as:
\begin{align}
    \phi_f^{t-1,t}=DCN_1(\phi_f^{t-1},o_1),\\
    \phi_f^{t-2,t}=DCN_2(\phi_f^{t-2},o_2).
\end{align}
Finally, the frame fusion layer integrates the features of $\phi_f^{t}$ with the aligned features of $\phi_f^{t-1,t}$ and $\phi_f^{t-2,t}$, producing the enhanced feature representation $\phi_f^{m,t}$ with inter-frame assistance, formulated as:
\begin{equation}
    \phi_f^{m,t}=FFL(\phi_f^{t},\phi_f^{t-1,t},\phi_f^{t-2,t}),
\end{equation}
where $FFL$ is the frame fusion layer network. The procedure of TCM at the high scale follows a similar paradigm with the medium scale, differing primarily in the operation of the offset layer. Specifically, the medium-scale processing learns absolute offsets directly, while the high-scale processing learns residual offsets relative to the medium-scale values, thereby enhancing offset precision and computational efficiency.

\subsection{Loss Functions}
We simulate real-world degradation by applying blur to the original visible videos and adding noise to the original infrared videos, while using original videos as anchors to constrain the network optimization. Initially, we employ the image-level fusion loss commonly used in multi-modal image fusion to constrain the perceptual quality of the fused images, which includes content loss and color loss:
\begin{align}
 \mathcal{L}_{con}&=\|\nabla V_f-\max(\nabla V_{vis},\nabla V_{ir})\|_1\nonumber\\
        &\qquad +\|V_f-\max(V_{vis},V_{ir})\|_1,\\
 \mathcal{L}_{cor}&=\|Cb_f-Cb_{vis}\|_1+\|Cr_f-Cr_{vis}\|_1.
\end{align}

Furthermore, as described in the previous sections, we enhance the feature representation capability of the model through feature distillation from existing foundation models VGG19 and Dinov2. This distillation loss can be formally expressed as:
\begin{align}
 \mathcal{L}_{d}&=\sum_{x}\|\phi_{x}^v-VGG19(V_{x})\|_2\nonumber\\
        &\qquad +\|\phi_{x}^s-Dinov2(V_{x})\|_2, x\in\{vis,ir\},
\end{align}
where $\phi_{x}^v$ and $\phi_{x}^s$ represent visual features and semantic features from $V_{x}$. To better enforce temporal consistency, we further design a temporal consistency loss. It ensures smooth inter-frame transitions by constraining the variation direction of the fused video to align with that of the original video, defined as:
\begin{equation}
    \begin{aligned}
        \mathcal{L}_{\text{t}}&=\sum_{x} ( \frac{1}{T-1} \sum_{t=1}^{T-1} ( 1 - \cos(V_f^t - V_f^{t-1}, V_{x}^t - V_{x}^{t-1}) ) ),
    \end{aligned}
\end{equation}
where $T$ denotes the number of frames. The final loss function can be concluded as:
\begin{equation}
 \mathcal{L}=\mathcal{L}_{con}+\lambda_1\mathcal{L}_{cor}+\lambda_2\mathcal{L}_d+\lambda_3\mathcal{L}_t,
\end{equation}
where $\lambda_1, \lambda_2, \lambda_3$ are hyperparameters, which are set to 2, 0.1, and 0.25, respectively. 
\section{Experiments}
In this section, we begin by detailing the implementation specifics and the datasets employed. Next, we introduce the proposed temporal consistency metrics. Subsequently, we compare the video fusion and video segmentation performance of our method with state-of-the-art approaches. Finally, we present ablation studies to validate the effectiveness of each design.

\subsection{Datasets and Implementation}
\textbf{Datasets.} We employ two infrared and visible video datasets for experiments, named M3SVD\footnote{\url{https://github.com/Linfeng-Tang/M3SVD}} and HDO\footnote{\url{https://github.com/xiehousheng/HDO}}. The M3SVD dataset includes 220 video sequences, with each sequence ranging from 100 to 1000 frames, while the HDO dataset comprises 26 video sequences, each containing between 150 to 200 frames.

\noindent\textbf{Implementation Details.} The TemCoCo network is trained for 20 epochs on M3SVD dataset. The learning rate is set to $5\times 10^{-4}$, decaying by a factor of 0.95 every 100 steps. Optimization is performed using the Adam optimizer, with a batch size of 3. The training videos are processed at a resolution of $640\times480$ with a temporal length of 10 frames, while the testing videos are set to the same resolution with a temporal length of 30 frames. Longer video sequences are constructed through overlapping concatenation to ensure temporal continuity between adjacent video slices. Since our focus is on processing video data with degradation types, such as blur in visible videos and stripe noise in infrared videos, we ensure a fair comparison by pre-processing the original degraded videos for baseline methods that do not consider degradation. Specifically, before performing the fusion, we employ state-of-the-art deblurring methods DSTNet~\cite{rao2024rethinking} tailored for visible videos and destriping methods MDIVDnet~\cite{cai2024exploring} designed for infrared videos, respectively. All experimented are processed on the PyTorch platform using a 2.0-GHz Intel Xeon Gold 6330 CPU and an NVIDIA A800 80GB PCIe GPU.

\subsection{Temporal Consistency Metrics}
In video fusion, it is essential to consider not only the quality of images but also the temporal consistency across video frames. Existing image fusion metrics fail to address this requirement. Therefore, we propose two novel metrics flowD and feaCD to evaluate inter-frame consistency, considering both the image-level and feature-level perspectives. We provide a comprehensive comparison between our proposed flowD, feaCD and existing video evaluation metrics (tOF~\cite{chu2018temporally} and LPIPS~\cite{zhang2018unreasonable}) in the supplementary material.

\noindent\textbf{flowD.} This metric predicts the next frame of the fused video by leveraging optical flow from the visible video and then compares it with the ground truth. The overall calculation is formulated as follows:
\vspace{-3pt}
\begin{equation}
    \begin{gathered}
 flowD(V_f,V_{vis})=\frac{1}{T-1}\sum_{t=1}^{T-1}\|warp^t-V_f^{t+1}\|_1,\\
 warp^t=warp(V_f^t,flow^t),
    \end{gathered}
    \label{eq:flowD}
\end{equation}
where $flow^t=flow(V_{vis}^t,V_{vis}^{t+1})$, and $flow$ is the network of Sea-raft~\cite{wang2024sea}, a state-of-the-art optical flow estimation method. Smaller flowD means more temporally consistent with the visible video.

\noindent\textbf{feaCD.} This metric evaluates the consistency of change direction between the fused video and the original video at the feature level. Specifically, video features are extracted using ResNet-18, and the cosine similarity between adjacent frames is computed to represent the change direction. The entire process can be formulated as follows:
\vspace{-5pt}
\begin{equation}
    \begin{gathered}
 \text{feaCD}(V_f, V_{vis}, V_{ir}) = \frac{1}{T-1}\sum_x \sum_{t=1}^{T-1} \| 1 - \cos_x \|_1, \\
 \cos_x = \cos(F_f^t - F_f^{t-1}, F_x^t - F_x^{t+1}),x\in\{vis,ir\},
    \end{gathered}
    \label{eq:feaCD}
\end{equation}
where $F_x^t=ResNet(V_x^t)$ and $F_f^t=ResNet(V_f^t)$. Smaller feaCD indicates more consistent directional changes with the original video.

\subsection{Video Fusion}
We conduct comparative experiments on video fusion. Due to the scarcity of existing video fusion methods, we select seven state-of-the-art image fusion algorithms and one video fusion algorithm: DATFuse~\cite{tang2023datfuse}, TGFuse~\cite{rao2023tgfuse}, DDFM~\cite{zhao2023ddfm}, CDDFuse~\cite{zhao2023cddfuse}, LRRNet~\cite{li2023lrrnet}, Diff-IF~\cite{yi2024diff}, MMIF-EMMA~\cite{zhao2024equivariant} and RCVS~\cite{xie2024rcvs}. These algorithms cover CNN-based methods, GAN-based methods, Transformer-based methods, and diffusion-based methods.

\begin{figure}[t]
    \centering
        \includegraphics[width=1\linewidth]{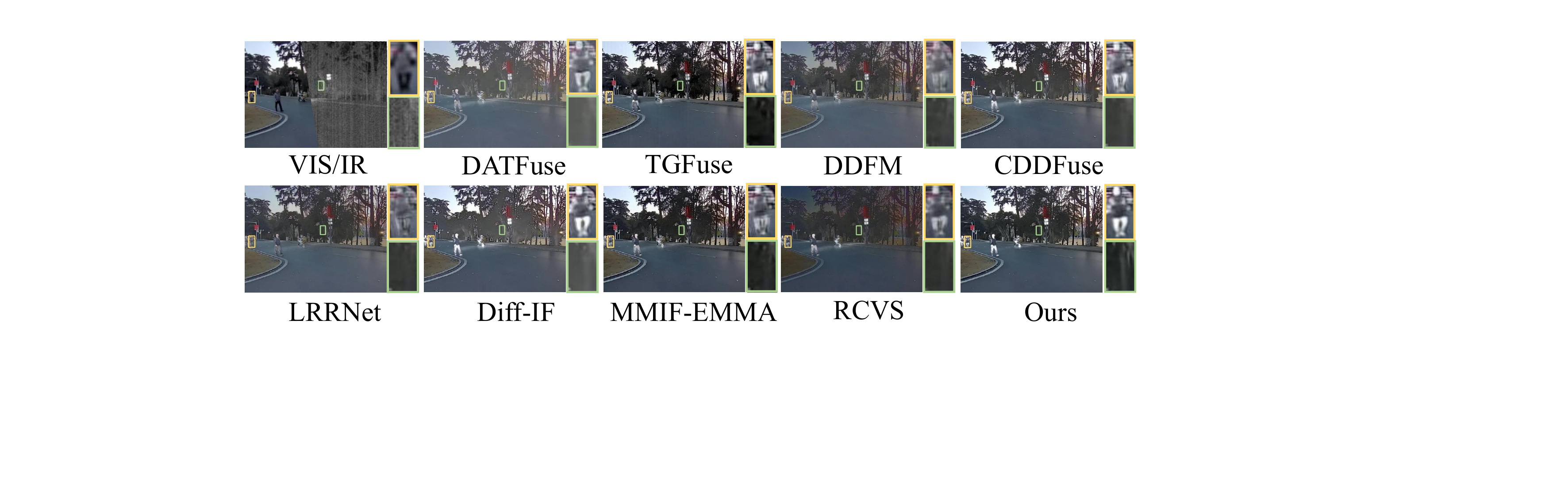}
        \vspace{-0.2in}
        \caption{Qualitative video fusion results on the M3SVD dataset.}
        \label{fig:MS3V_1}
\end{figure}

\begin{figure}[t]
    \centering
        \includegraphics[width=1\linewidth]{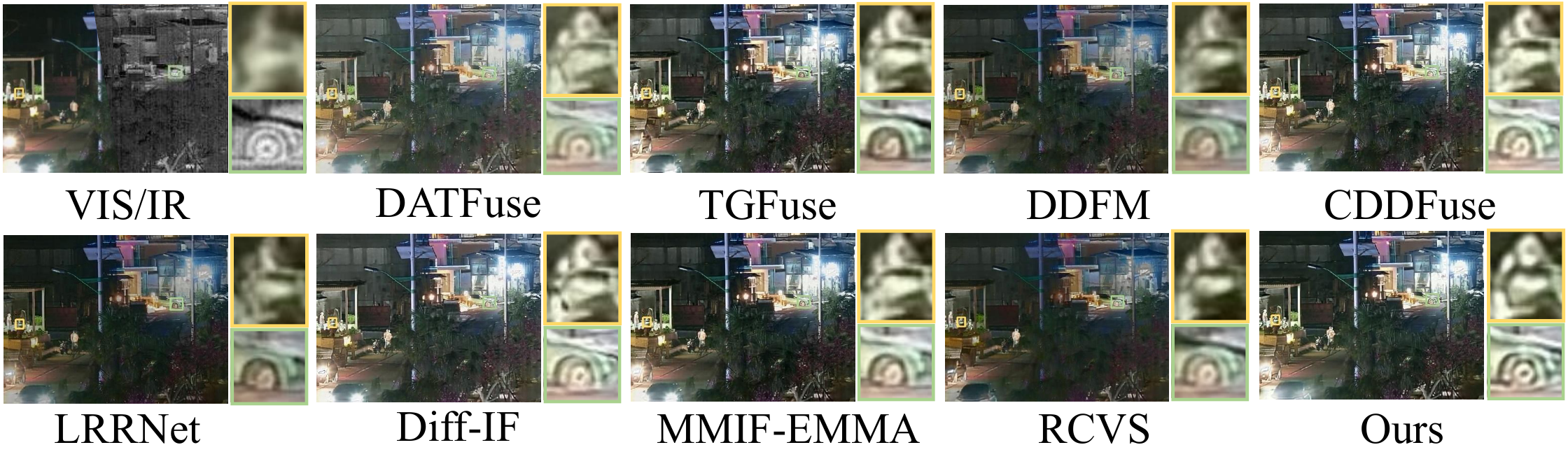}
        \vspace{-0.2in}
        \caption{Qualitative video fusion results on the HDO dataset.}
        \label{fig:HDO_1}
\end{figure}

\begin{figure}[t]
    \centering
        \includegraphics[width=1\linewidth]{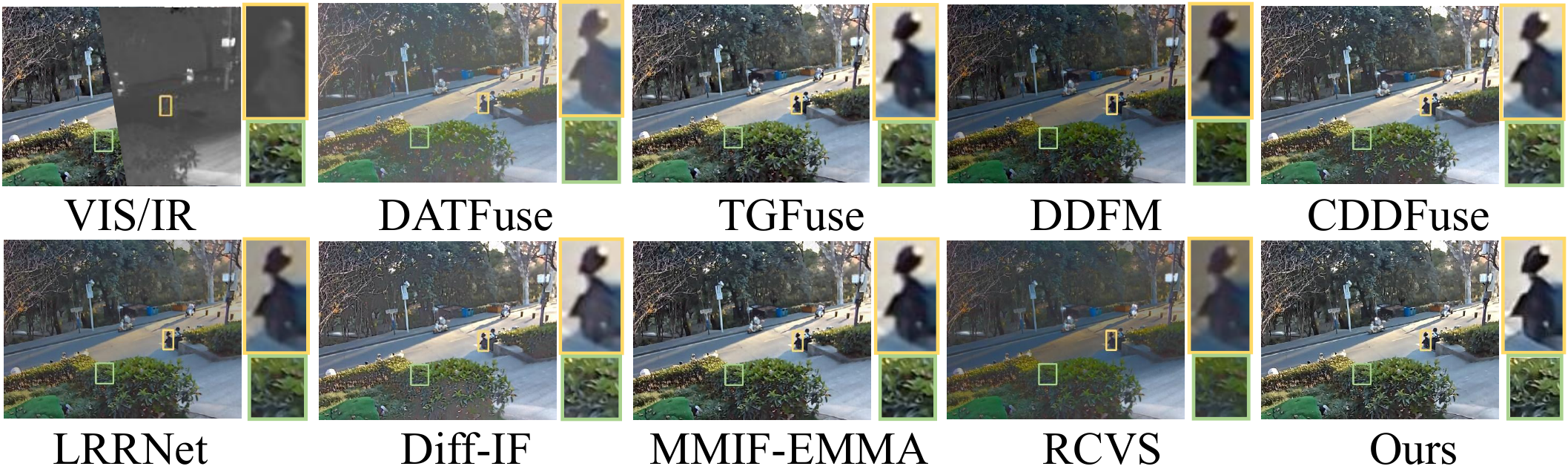}
        \vspace{-0.2in}
        \caption{Qualitative video fusion results with real-world degradation.}
        \label{fig:true_blur}
\end{figure}

\begin{figure}[t]
    \centering
        \includegraphics[width=1\linewidth]{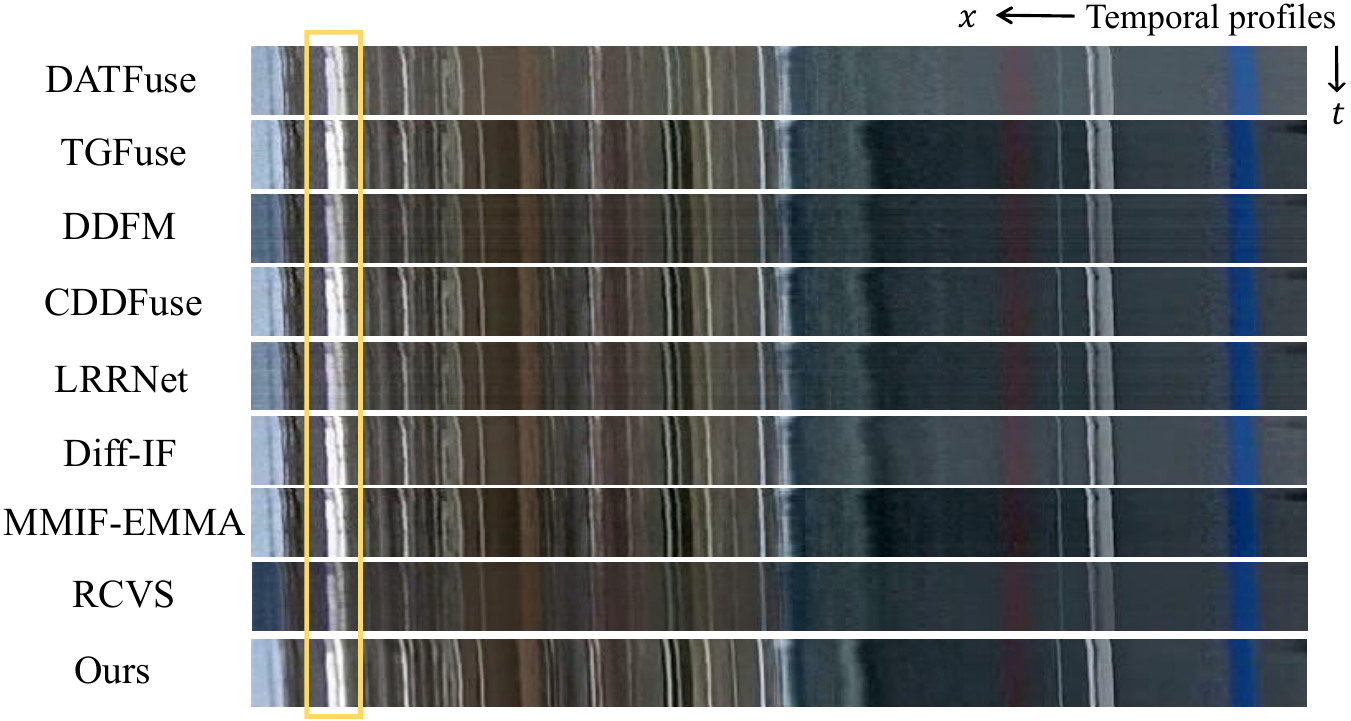}
        \vspace{-0.2in}
        \caption{The results of the 150th row from the 10th to the 50th frame extracted along the temporal dimension.}
        \label{fig:temporal_fusion}
\end{figure}
We represent qualitative video fusion results on the M3SVD dataset and the HDO dataset in Figure~\ref{fig:MS3V_1} and Figure~\ref{fig:HDO_1}, respectively. We present the enlarged results of the corresponding colored boxes on the right side of each result image. In Figure~\ref{fig:MS3V_1}, we first focus on the magnified results within the yellow bounding box. Our method demonstrates clearer and more salient targets. Other methods show blurred details and fail to effectively highlight thermal targets. Furthermore, as shown in the green bounding box, only our method successfully preserves the fine details of the tree branches, whereas all other methods lose this critical information. Similarly, in the yellow boxes of Figure~\ref{fig:HDO_1}, only our approach successfully preserves the contour and details of the `person pouring tea'. In the green boxes, our method also demonstrates the sharpest details, particularly in the wheel of the car. Furthermore, we present experimental results on real-world degraded scenarios, as illustrated in Figure~\ref{fig:true_blur}. It can be observed that the original visible video suffers from significant blurring. Only our method demonstrates robustness to such degradation, producing clearer fusion results. The enhanced details are particularly evident in the zoomed-in regions.

\begin{table}[]
\centering
\caption{Quantitative video fusion results on the M3SVD dataset. \textbf{Bold} is the best.}
\label{tab:M3SVD}
\resizebox{\linewidth}{!}{
    \begin{tabular}{c|cccc}
    \toprule
    \multicolumn{1}{c|}{M3SVD} & MI$\uparrow$    & SD$\uparrow$        & flowD$\downarrow$     & feaCD$\downarrow$           \\ \midrule
    DATFuse               & 1.9902          & 32.0269          & 4.0048          & 1.8709          \\
    TGFuse                & 2.2379          & 45.9510          & 4.3461          & 1.8338          \\
    DDFM                  & 1.6483          & 26.4525          & 5.1480          & 1.9416          \\
    CDDFuse               & 2.1694          & 45.4648          & 4.5214          & 1.9183          \\
    LRRNet                & 2.0207          & 32.9157          & 5.0990          & 1.9417          \\
    Diff-IF               & 2.2491          & 42.4253          & 3.7864          & 1.8346          \\
    MMIF-EMMA             & 2.3140          & 47.0037          & 4.3144          & 1.8598          \\
    RCVS            & 1.1548          & 29.2992         & 3.8512          & 1.8160          \\\hline\\[-1em]
    Ours                  & \textbf{2.4219} & \textbf{50.5400} & \textbf{3.2847} & \textbf{1.8049} \\ \bottomrule
    \end{tabular}}
\end{table}

\begin{table}[]
\centering
\caption{Quantitative video fusion results on the HDO dataset. \textbf{Bold} is the best.}
\label{tab:HDO}
\resizebox{\linewidth}{!}{
    \begin{tabular}{c|cccc}
    \toprule
    \multicolumn{1}{c|}{HDO} & MI$\uparrow$    & SD$\uparrow$        & flowD$\downarrow$     & feaCD$\downarrow$           \\ \midrule
    DATFuse               & 1.1115          & 48.9679          & 8.0831          & 1.9791          \\
    TGFuse                & 1.0155          & 58.4364          & 8.6420          & 1.9510          \\
    DDFM                  & 1.0265          & 43.5832          & 6.9643          & 1.9714          \\
    CDDFuse               & 1.1396          & 60.3634          & 8.3442          & 1.9460          \\
    LRRNet                & 1.0616          & 43.3673          & 7.4598          & 1.9523          \\
    Diff-IF               & 1.1397          & 57.4811          & 8.0450          & 1.9587          \\
    MMIF-EMMA             & 1.1668          & \textbf{60.4379}          & 7.9598          & 1.9565          \\
    RCVS            & 0.9617         & 44.2517         & 7.1108          & 1.9657          \\\hline\\[-1em]
    Ours                  & \textbf{1.3373} & {59.4020} & \textbf{5.1147} & \textbf{1.7336} \\ \bottomrule
    \end{tabular}}
\end{table}
To visualize the advantages of our method in processing video sequences, we present the results of a specific row concatenated along the temporal dimension in Figure~\ref{fig:temporal_fusion}. The results demonstrate that our method achieves smoother transitions with minimal fluctuations compared to other approaches, highlighting its superior temporal consistency.

Finally, we report quantitative results on the M3SVD and HDO datasets, as shown in Table~\ref{tab:M3SVD} and Table~\ref{tab:HDO}. Four metrics are used: MI (reference-based) and SD (no-reference) for image fusion quality, and flowD and feaCD for temporal consistency. TemCoCo outperforms state-of-the-art methods in both aspects, with notable gains in MI under degraded conditions, highlighting its robustness. Furthermore, our method achieves superior temporal consistency, validating the effectiveness of the temporal designs.


\begin{figure}[t]
    \centering
        \includegraphics[width=1\linewidth]{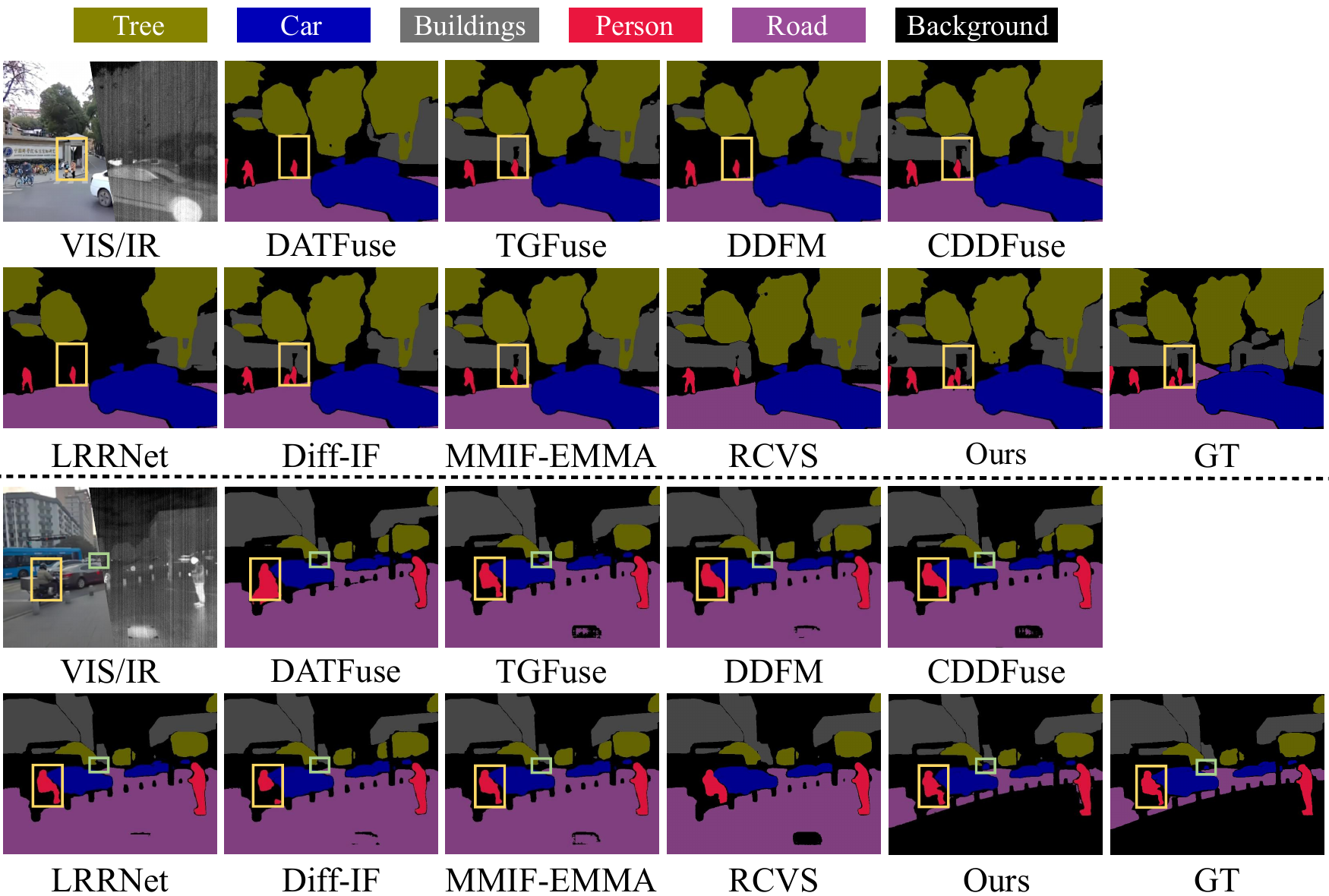}
        \vspace{-0.2in}
        \caption{Two examples of video segmentation results.}
        \label{fig:seg_2}
\end{figure}

\begin{figure}[t]
    \centering
        \includegraphics[width=1\linewidth]{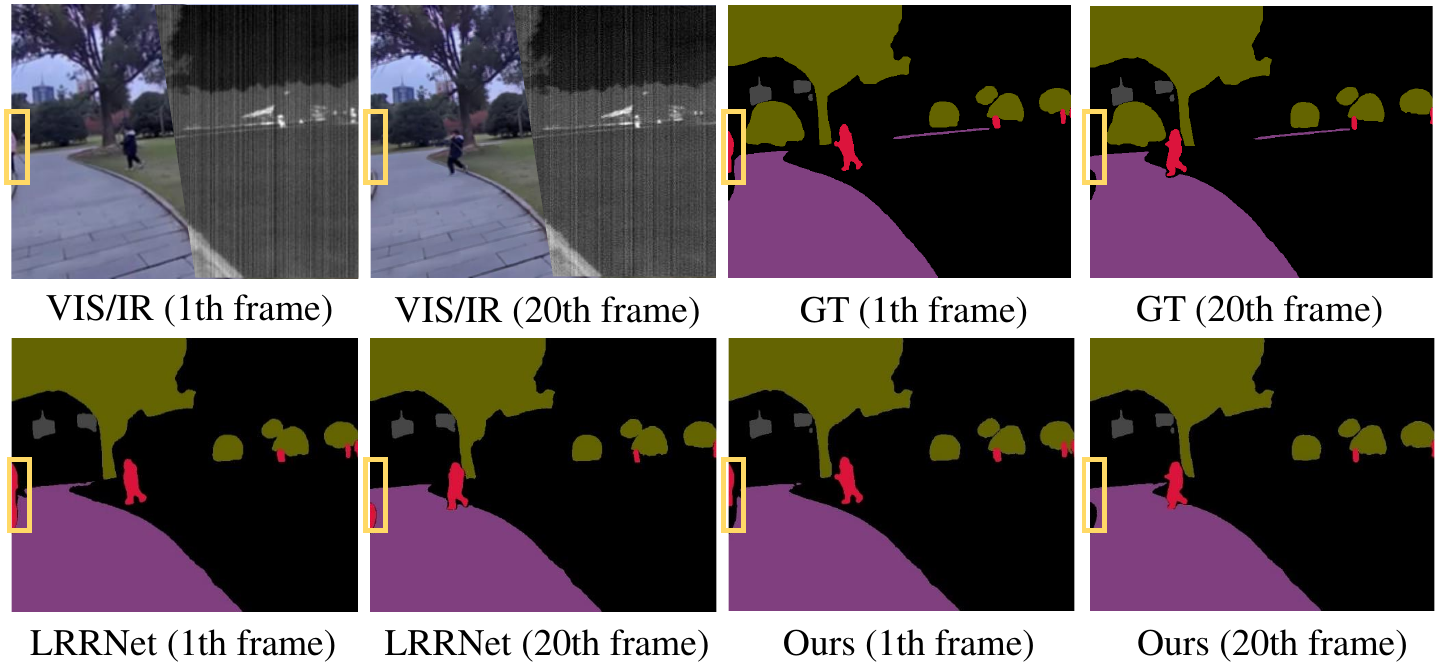}
        \vspace{-0.2in}
        \caption{Temporal video segmentation results.}
        \label{fig:temporal_seg}
\end{figure}

\subsection{Video Segmentation}
In this part, we present the performance of our proposed TemCoCo compared to other algorithms on the video segmentation task. Specifically, we employ the GroundedSAM2~\cite{ren2024grounding, ravi2024sam2segmentimages} model with text prompts "Road, Person, Tree, Buildings, Car" to perform video segmentation. Due to the lack of a dedicated video segmentation dataset for infrared-visible videos, we manually annotate five video sequences, each comprising 50 frames, with the assistance of SAM2~\cite{ravi2024sam2segmentimages}. The annotations include six classes: road, person, tree, buildings, car, and background.

Figure~\ref{fig:seg_2} presents the qualitative segmentation results of two annotated videos alongside their ground-truth labels. In the first example, our method shows superior ability in distinguishing buildings from the background, achieving the highest consistency with the ground truth. In contrast, methods such as DATFuse, DDFM, and LRRNet fail to retain significant parts of the buildings. Meanwhile, TGFuse, CDDFuse, MMIF-EMMA and RCVS are unable to accurately segment the person. Although Diff-IF achieves relatively complete segmentation, it still misses the door part of the building. Similarly, in the second example, only our method and TGFuse successfully segment the car within the green bounding box, and our method is the sole approach that effectively differentiates the road from the background.

Furthermore, in Figure~\ref{fig:temporal_seg}, we present an example to highlight the unique challenges of video segmentation compared to image segmentation. Theoretically, only the first frame of video segmentation relies entirely on detection results, while subsequent frames depend on the segmentation outcomes of preceding frames. As shown in the figure, in the 1st and 20th frames, the person on the far left gradually exits the scene. Due to an initial mis-detection by LRRNet, which incorrectly includes part of the step as the person, the 20th frame erroneously continues to classify the step as part of the person. In contrast, our method correctly distinguishes the person in the first frame and maintains accurate segmentation even after the person leaves.

Finally, in Table~\ref{tab:IoU} and Table~\ref{tab:Acc}, we report the average segmentation results on the annotated data, including the segmentation accuracy (Acc) and intersection over union (IoU) for each category, as well as the overall mean Acc (mAcc) and mean IoU (mIoU) across all classes. Our method achieves optimal performance in terms of both Acc and IoU across most categories, further demonstrating that our approach effectively integrates semantic information and is applicable to downstream tasks.

\begin{table}[]
\addtolength{\tabcolsep}{-3pt}
\centering
\caption{Quantitative IoU results of video segmentation.}
\label{tab:IoU}
\resizebox{\linewidth}{!}{
    \begin{tabular}{c|cccccc|c}
    \toprule
    \multicolumn{1}{c|}{IoU} & Road   & Person  & Tree  & Build.  & Car  & Back.  & {mIoU}       \\ \midrule
    DAT.               & 0.724  & 0.733  & 0.712  & 0.686  & 0.754  & 0.651  & 0.710    \\
    TGF.                & 0.736  & 0.779  & \textbf{0.906}  & 0.839  & 0.763  & 0.724  & 0.791      \\
    DDF.                  & 0.739  & 0.759  & 0.876  & 0.811  & 0.756  & 0.716  & 0.774      \\
    CDD.               & 0.736  & 0.753  & 0.735  & 0.882  & 0.766  & 0.806  & 0.763     \\
    LRR.                & 0.733  & 0.738  & 0.829  & 0.807  & 0.754  & 0.672  & 0.755      \\
    Dif.               & 0.730  & 0.734  & 0.824  & 0.811  & 0.882  & 0.765  & 0.712          \\
    MMI.             & 0.738  & 0.778  & 0.726  & 0.839  & 0.761  & 0.691  & 0.755 \\
    RCV.             & 0.738  & 0.732  & 0.868  & 0.795  & 0.755  & 0.706  & 0.766 \\ \hline\\[-1em]
    Ours                  & \textbf{0.846} & \textbf{0.825} & {0.901} & \textbf{0.904} & \textbf{0.900} & \textbf{0.847} & \textbf{0.870}\\
    nodino & 0.845  & 0.710  & 0.748  & 0.807  & 0.695  & 0.784  & 0.765      \\ \bottomrule
    \end{tabular}}
\end{table}

\begin{table}[]
\addtolength{\tabcolsep}{-3pt}
\caption{Quantitative Acc results of video segmentation.}
\label{tab:Acc}
\centering
\resizebox{\linewidth}{!}{
    \begin{tabular}{c|cccccc|c}
    \toprule
    \multicolumn{1}{c|}{Acc} & Road   & Person  & Tree  & Build.  & Car  & Back.  & {mAcc}       \\ \midrule
    DAT.               & 0.925  & 0.837  & 0.754  & 0.805  & 0.760  & 0.780  & 0.810    \\
    TGF.                & 0.957  & 0.826  & \textbf{0.932}  & 0.877  & 0.771  & 0.799  & 0.860      \\
    DDF.                  & \textbf{0.959}  & 0.835  & 0.918  & 0.853  & 0.761  & 0.810  & 0.856      \\
    CDD.               & 0.955  & 0.820  & 0.773  & {0.934}  & 0.771  & 0.806  & 0.843     \\
    LRR.                & 0.950  & 0.822  & 0.849  & 0.847  & 0.767  & 0.811  & 0.841      \\
    Dif.               & 0.949  & 0.814  & 0.863  & 0.935  & 0.774  & 0.798  & 0.856          \\
    MMI.             & 0.956  & 0.838  & 0.767  & 0.877  & 0.774  & 0.798  & 0.835      \\
    RCV.             & 0.958  & 0.827  & 0.903  & 0.933  & 0.760  & 0.775  & 0.860 \\ \hline\\[-1em]
    Ours                  & {0.957} & \textbf{0.862} & {0.916} & \textbf{0.942} & \textbf{0.914} & \textbf{0.920} & \textbf{0.919}\\
    nodino & 0.942  & 0.765  & 0.783  & 0.850  & 0.743  & 0.894  & 0.830      \\ \bottomrule
    \end{tabular}}
\end{table}

\begin{table}[]
\centering
\caption{Quantitative video fusion results of ablation studies.}
\label{tab:ablation}
\resizebox{\linewidth}{!}{
    \begin{tabular}{c|cccc}
    \toprule
    \multicolumn{1}{c|}{M3SVD} & MI$\uparrow$    & SD$\uparrow$        & flowD$\downarrow$     & feaCD$\downarrow$           \\ \midrule
    wo\_TLoss               & 2.0142          & 50.4149          & 5.7999          & 1.9374          \\
    wo\_TAttn                & 2.1704          & 50.4297          & 4.4849          & 1.8645          \\
    no\_VSAttn                 & 2.1392          & 50.2262          & 4.1371          & 1.8273          \\
    no\_DCN               & 2.2125          & 50.2244          & 3.2879          & 1.8160          \\
    no\_Dino                & 2.2591          & 50.1774           & 3.2856          & 1.8075          \\
    no\_Dino\_VGG               & 2.1612          & 49.9946         & 3.2936          & 1.8124          \\
    Conv2d             & 2.3400          & 50.2530          & 3.9472         & 1.8236          \\\hline\\[-1em]
    Ours                  & \textbf{2.4219} & \textbf{50.5400} & \textbf{3.2847} & \textbf{1.8049} \\ \bottomrule
    \end{tabular}}
\end{table}

\subsection{Ablation Studies}
We conduct extensive ablation experiments to comprehensively validate the effectiveness of each proposed strategy or module, including the following configurations: \textbf{no\_TLoss}: Removal of the temporal consistency loss. \textbf{no\_TAttn}: Removal of the temporal attention mechanism. \textbf{no\_VSAttn}: Removal of the visual-semantic attention mechanism. \textbf{no\_DCN}: Replacement of the DCN module with a module that only associates with the current frame. \textbf{no\_Dino}: Removal of the Dinov2 distillation guidance. \textbf{no\_Dino\_VGG}: Removal of both Dinov2 and VGG19 distillation guidance. \textbf{Conv2d}: Replacement of all Conv3d layers with Conv2d layers.

\begin{figure}[t]
    \centering
        \includegraphics[width=1\linewidth]{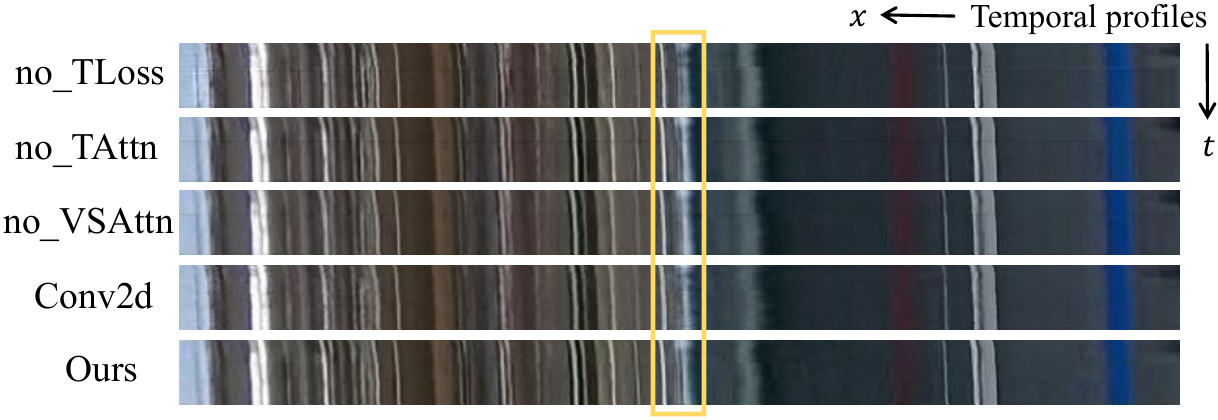}
        \vspace{-0.2in}
        \caption{Temporal results of ablation studies.}
        \label{fig:ablation_temporal}
\end{figure}

\begin{figure}[t]
    \centering
        \includegraphics[width=1\linewidth]{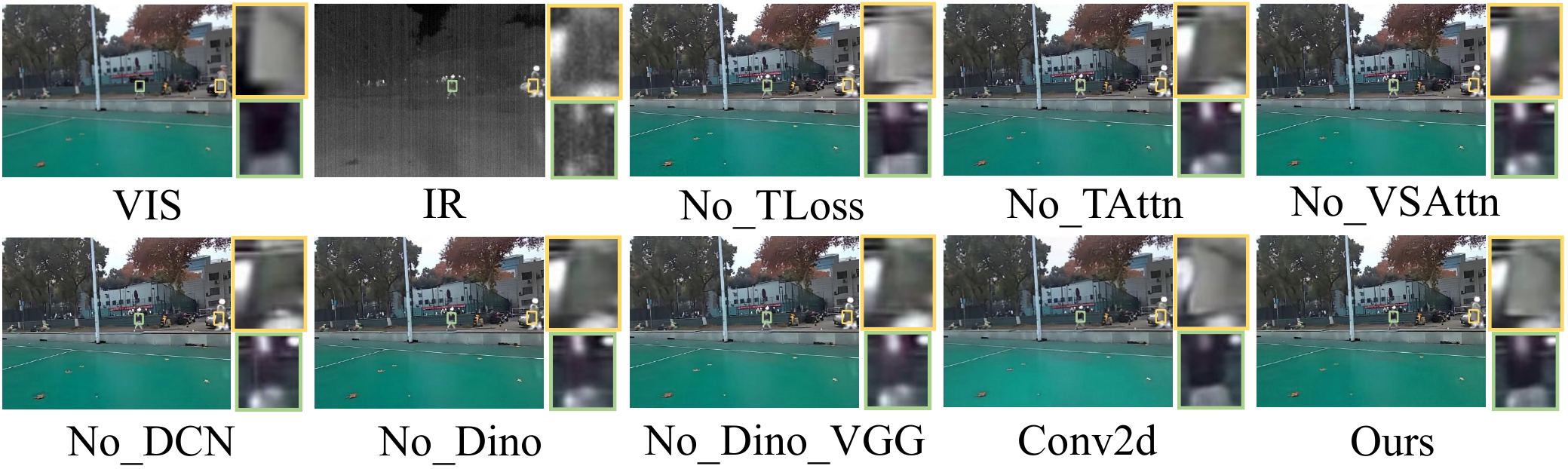}
        \vspace{-0.2in}
        \caption{Qualitative video fusion results of ablation studies.}
        \label{fig:ablation}
\end{figure}

We present the quantitative results for all test videos in Table~\ref{tab:ablation}. All ablation variants show a significant drop in image fusion metrics, especially the supervised metric MI, indicating the positive contribution of each module to weak information recovery. However, only a subset of experiments demonstrates a notable impact on temporal consistency, such as no\_TLoss, no\_TAttn, no\_VSAttn, and Conv2d, which aligns well with our design objectives. To further investigate these findings, Figure~\ref{fig:ablation_temporal} presents qualitative comparisons of inter-frame smoothness, where clear transition artifacts emerge in the first three configurations. Given the MI degradation across all variants, we also include visual comparisons in Figure~\ref{fig:ablation}, where our method produces the clearest textures and most detailed contours, further validating the effectiveness of these components. Finally, we present the impact of removing Dinov2 on segmentation results in the last row of Tables~\ref{tab:IoU} and~\ref{tab:Acc}. The results indicate that when the semantic guidance of Dinov2 is eliminated, our method performs comparably to other image-based fusion methods. This is because, although our method excels in temporal consistency, the video segmentation results remain highly dependent on the detection accuracy of the first frame. This finding further validates that the introduction of Dinov2 indeed provides significant semantic information enhancement for the segmentation task, thereby improving the overall segmentation performance.

\begin{table}[t]
    \centering
    \scriptsize
    \setlength{\tabcolsep}{2pt}
\centering
\caption{Efficiency comparison (FLOPs (G), Time (s)).}
\label{tab:efficiency}
\resizebox{\linewidth}{!}{
    \begin{tabular}{lcccccccccc}
    \toprule
    \textbf{Methods} & DAT. & TGF. & DDF. & CDD. & LRR. & Dif. & MMI. & RCV. & Ours-2d & Ours \\
    \midrule
    \textbf{FLOPs} & 5.6 & 95.7 & 5220.5 & 364.2 & 14.2 & 86.9 & 41.5 & \textbf{0.05} & 442.4 & 505.6 \\
    \textbf{Time}  & 0.061 & 0.143 & 34.494 & 0.511 & 0.162 & 4.233 & 0.056 & \textbf{0.009} & 0.086 & 0.109 \\
    \bottomrule
    \end{tabular}}
    \vspace{-0.18in}
\end{table}

\subsection{Efficiency Comparison}
We also give the efficiency comparison results in Table~\ref{tab:efficiency}. As multi-frame processing shares computation across frames, our TemCoCo remain moderate inference time and acceptable FLOPs. Importantly, our model adopts 3D separable convolutions to effectively capture temporal information with minimal additional overhead, enabling a notable performance gain at relatively low cost. As shown in the table, the computational overhead of using 3D convolutions over 2D counterparts is minor, while the performance improvement shown in Figure~\ref{fig:ablation} is significant, highlighting the favorable trade-off and necessity of our 3D design.

\section{Conclusion}
This paper presents the first robust multi-modal video fusion framework that explicitly models temporal consistency, while simultaneously addressing the challenges of visual fidelity and semantic accuracy. First, a visual-semantic interaction module, enhanced with Dinov2 and VGG19 distillation, achieves the explicit integration of semantic guidance into video fusion. Second, the temporal cooperative module leverages inter-frame dependencies to recover weak information, pioneering the joint handling of video degradation and fusion in a unified framework. Third, the temporal-enhanced mechanism and the dedicated temporal loss constitute the first explicit design for modeling and enforcing temporal consistency in multi-modal video fusion. Additionally, the two newly introduced metrics provide a more comprehensive and precise quantitative tool for evaluating the temporal consistency. Extensive video experimental results on public datasets demonstrate the superiority. We believe this work opens new avenues for future research in video fusion and related fields.

\section*{Acknowledgements}
This work was supported by the National Natural Science Foundation of China (62276192), and the Fund of National Key Laboratory of Multispectral Information Intelligent Processing Technology (61421132302).

{
    \small
    \bibliographystyle{ieeenat_fullname}
    \bibliography{main}
}

\end{document}